\documentclass[conference]{IEEEtran}
\IEEEoverridecommandlockouts

\usepackage[nocompress]{cite}

\usepackage{amsmath,amssymb,amsfonts}
\usepackage{algorithmic}
\usepackage{graphicx}
\usepackage{textcomp}
\usepackage{xcolor}
\usepackage{url}
\usepackage{bm}
\usepackage{multirow}
\usepackage{booktabs}
\usepackage{placeins}
\usepackage{siunitx}
\def\BibTeX{{\rm B\kern-.05em{\sc i\kern-.025em b}\kern-.08em
    T\kern-.1667em\lower.7ex\hbox{E}\kern-.125emX}}

\newcommand\copyrighttext{
    \footnotesize \copyright{ }2026 IEEE. Personal use of this material is permitted. Permission from IEEE must be obtained for all other uses, in any current or future media, including reprinting/republishing this material for advertising or promotional purposes, creating new collective works, for resale or redistribution to servers or lists, or reuse of any copyrighted component of this work in other works.}
\newcommand\copyrightnotice{
    \begin{tikzpicture}[remember picture,overlay]
    \node[anchor=south,yshift=15pt,xshift=0pt] at (current page.south) {\parbox{\dimexpr\textwidth-\fboxsep-\fboxrule\relax}{\copyrighttext}};
    \end{tikzpicture}
}

\usepackage{pict2e}
\newcommand{\colorcircle}[1]{
  {\unitlength=1ex
   \begin{picture}(1.5,1)
     \linethickness{0.1ex}
     \put(0.7,0.7){\color{#1}\circle*{1.6}}
     \put(0.7,0.7){\color{black}\circle{1.6}}
   \end{picture}}
}

\definecolor{rwthblue}{RGB}{0,84,159}
\definecolor{rwthlightblue}{RGB}{142,186,229}
\definecolor{rwthblack}{RGB}{0,0,0}
\definecolor{rwthgrey}{RGB}{100,101,103}
\definecolor{rwthpetrol}{RGB}{0,97,101}
\definecolor{rwthturquoise}{RGB}{0,152,161}
\definecolor{rwthgreen}{RGB}{87,171,39}
\definecolor{rwthmaygreen}{RGB}{189,205,0}
\definecolor{rwthyellow}{RGB}{255,237,0}
\definecolor{rwthorange}{RGB}{246,168,0}
\definecolor{rwthred}{RGB}{204,7,30}
\definecolor{rwthmagenta}{RGB}{227,0,102}
\definecolor{rwthbordeaux}{RGB}{161,16,53}
\definecolor{rwthviolet}{RGB}{97,33,88}
\definecolor{rwthpurple}{RGB}{122,111,172}

\usepackage{etex}
\usepackage{booktabs}
\usepackage{pgfplots}
\pgfplotsset{compat=1.15}
\usepackage{tikz}
\usetikzlibrary{positioning}

\begin{document}
\title{Robust Fusion of Object-Level V2X for Learned 3D Object Detection}

\author{
    Lukas Ostendorf$^{1}$,
    Lennart Reiher$^{1}$,
    Onn Haran$^{2}$, 
    Lutz Eckstein$^{\dagger}$
    \thanks{$^{1}$ The authors are with the Institute for Automotive Engineering, RWTH Aachen University, 52074 Aachen, Germany. \texttt{{firstname.lastname}@ika.rwth-aachen.de}}
    \thanks{$^{2}$ The author is with Qualcomm Technologies, Inc., Kfar-Netter, Israel. This research has been funded by Qualcomm.}
    \thanks{$^{\dagger}$ Lutz Eckstein is head of the Institute for Automotive Engineering (ika).}
}
\maketitle

\copyrightnotice

\begin{abstract}
Perception for automated driving is largely based on onboard environmental sensors, such as cameras and radar, which are cost-effective but limited by line-of-sight and field-of-view constraints. These inherent limitations may cause onboard perception to fail under occlusions or poor visibility conditions. In parallel, cooperative awareness via vehicle-to-everything (V2X) communication is becoming increasingly available, enabling vehicles and infrastructure to share their own state as object-level information that complements onboard perception. In this work, we study how such V2X information can be integrated into 3D object detection and how robust the resulting system is to realistic V2X imperfections.
Using the \textit{nuScenes} dataset, we emulate object-level cooperative awareness messages from ground truth, injecting controlled noise and object dropout to mimic real-world conditions such as latency, localization errors, and low V2X penetration rates. We convert these messages into a dedicated bird’s-eye view (BEV) input and fuse them into a \textit{BEVFusion}-style detector. Our results demonstrate that while object-level cooperative information can substantially improve detection performance, achieving an NDS of 0.80 under favorable conditions, models trained on idealized data become fragile and over-reliant on V2X. Conversely, our proposed noise-aware training strategy, coupled with explicit confidence encoding, enhances robustness, maintaining performance gains even under severe noise and reduced V2X penetration.
\end{abstract}


\section{Introduction}
Road traffic accidents remain a leading cause of death and injury worldwide \cite{WHO_road_safety}. Automated and autonomous driving systems aim to reduce the number of accidents, relying on perception systems that must be accurate and reliable under diverse conditions. In current series-production vehicles, perception is typically based on onboard environment sensors such as cameras and radar. Cameras are low-cost and provide rich semantic information but suffer from limited field-of-view and failures in adverse weather. Radar is more robust to weather and provides direct velocity measurements but has limited angular resolution. LiDAR offers dense, accurate 3D measurements, but remains prohibitively costly for most series-production vehicles. As a result, purely onboard environment sensing faces fundamental limitations in complex traffic scenes.

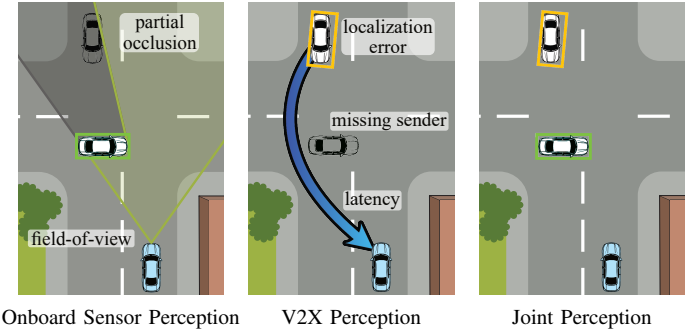
\begin{figure}[t]
    \centering
    \begin{tikzpicture}
        \node[inner sep=0pt] (sensor) at (0, 0) {\includegraphics[width=0.15\textwidth]{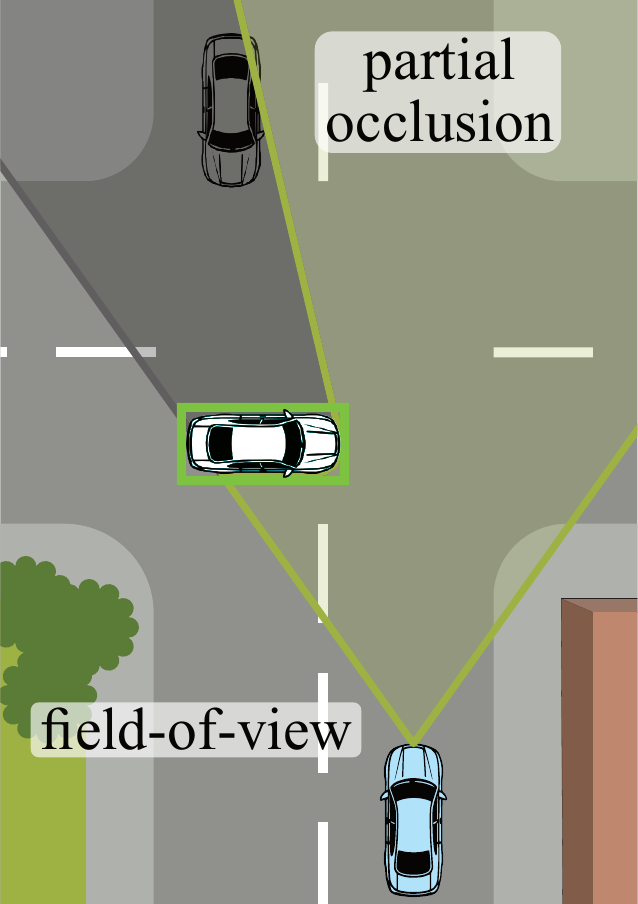}};
        \node[below=2pt of sensor, font=\footnotesize] {Onboard Sensor Perception};
        
        \node[inner sep=0pt, right=9pt of sensor] (v2x) {\includegraphics[width=0.15\textwidth]{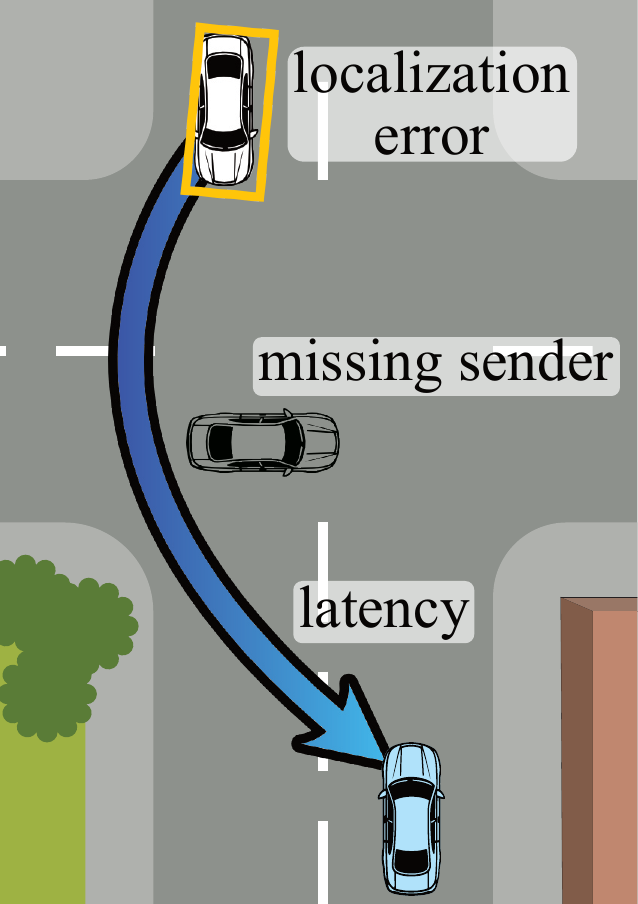}};
        \node[below=2pt of v2x, font=\footnotesize] {V2X Perception};

        \node[inner sep=0pt, right=9pt of v2x] (joint) {\includegraphics[width=0.15\textwidth]{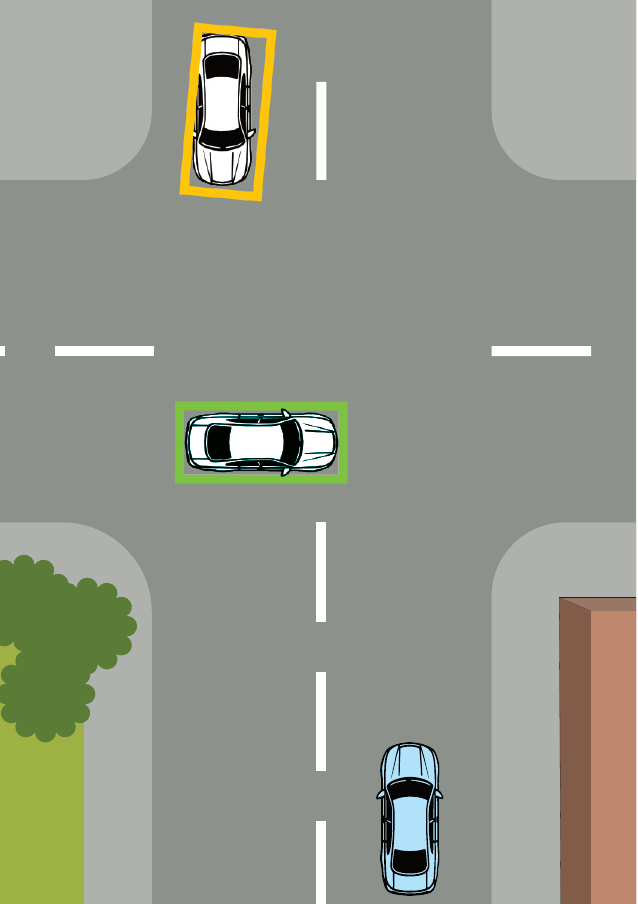}};
        \node[below=2pt of joint, font=\footnotesize] {Joint Perception};
    \end{tikzpicture}
    \vspace{-0.5cm}
    \caption{Comparison of perception modalities: (left) onboard sensors with limited field-of-view and occlusions, (center) V2X that reveals occluded actors but suffers localization error and latency, and (right) joint perception that combines both for more robust scene understanding.}
    \label{fig:intro_joint_perception}
\end{figure}

As illustrated in Figure~\ref{fig:intro_joint_perception}, vehicle-to-everything (V2X) communication provides complementary information by enabling vehicles and infrastructure to exchange self-localized states. Thereby, it extends sensing horizons and exposes participants that would otherwise remain hidden. Object-level data in V2X messages can, in principle, cover blind spots of onboard environment sensors and provide semantic cues. In practice, V2X messages suffer from imperfections. Positioning based on a Global Navigation Satellite System (GNSS) can be off by several meters in urban canyons~\cite{KaplanHegarty2017,Groves2011ShadowMatching}, and end-to-end delivery often incurs tens to hundreds of milliseconds of latency~\cite{CollPerales2023E2ELatency}. In bird’s-eye view (BEV), these issues appear as spatial misalignment and temporal lag.

Na\"{\i}vely overlaying stale or mislocalized V2X object information on top of camera or radar detections risks false associations, hallucinated objects, and brittle performance. To use V2X as an additional input for 3D object detection, it is necessary to understand how to integrate this information into modern multi-modal networks and how sensitive such integration is to localization errors and low V2X penetration rates. In practice, not all vehicles transmit V2X messages, and some object classes may never participate (e.g., cones, barriers, pedestrians), so robust integration strategies must handle partial or incomplete V2X information.

In this work, we focus on object-level V2X information that provides object states (position, orientation, velocity) from remote vehicles and infrastructure. We use ground-truth annotations from the \textit{nuScenes} dataset~\cite{caesar2020nuscenes} to emulate such V2X information under controlled noise and object dropout. Our emulation explicitly models latency and localization errors via object-level and frame-level rotation and translation noise. It also incorporates object-level and class-level dropout to account for vehicles that lack V2X capability and object classes that do not participate in V2X communication. We integrate the V2X-derived BEV input into a \textit{BEVFusion}-style~\cite{bevfusion} architecture using a dedicated V2X encoder and gating-based fusion mechanism, for both camera-only and radar+camera configurations. This design allows us to systematically study how V2X information improves detection and how performance degrades as V2X becomes noisy, incomplete, or sparse.

We address three key questions:
\begin{itemize}
    \item \textbf{How can V2X information be incorporated into 3D object detection frameworks?} We propose a controllable V2X emulation pipeline with a dedicated encoder and gating-based fusion for \textit{BEVFusion}~\cite{bevfusion}. 
    \item \textbf{How robust is such integration to V2X noise and limited V2X penetration rates?} We conduct systematic experiments on \textit{nuScenes}~\cite{caesar2020nuscenes} and demonstrate that noise-aware training with confidence encoding improves robustness. 
    \item \textbf{How does performance compare to camera-only and radar baselines across varying V2X quality?} We show that appropriate training strategies maintain gains even under severe imperfections.
\end{itemize}

\section{Related Work}

We review V2X communication standards and their integration into learning-based perception, then discuss 3D object detection methods that form the foundation of our approach.

\subsection{V2X Communication}

Vehicle-to-Everything communication enables vehicles, infrastructure, and other road users to share information wirelessly. In Europe, it is standardized by the European Telecommunications Standards Institute (ETSI). Within this framework, Cooperative Awareness Messages (CAMs) broadcast a sender’s own state: position, motion, and basic status, so nearby participants know where it is and what it is doing. Collective Perception Messages (CPMs) build on this by sharing the objects detected in the sender’s surroundings, enabling others to benefit from its sensing capabilities.

A key challenge when using V2X data as auxiliary perception input is the spatiotemporal inaccuracy of transmitted information. GNSS-based positioning suffers from satellite blockage, multipath, and atmospheric effects that introduce meter-level and time-varying errors~\cite{KaplanHegarty2017,Groves2011ShadowMatching}. Message timeliness is further degraded by channel contention and processing delays, creating latency and jitter that misalign received states with the local perception frame~\cite{Emara2018MEC,CollPerales2023E2ELatency}. Recent datasets, such as V2AIX~\cite{Kueppers2024V2AIX}, empirically characterize these real-world effects, highlighting the need for fusion strategies that can reliably accommodate uncertain and delayed information.

\subsection{V2X in Learning-Based Perception}
Recent work demonstrates that V2X information can substantially enhance learning-based perception. Integration methods span a spectrum from early fusion of exchanged point clouds~\cite{Chen2019COOPER} to mid-level sharing of latent features~\cite{Chen2019FCOOPER, Xu2022V2XViT} and late fusion of object lists~\cite{Yuan2022_FPVRCNN}, enabling agents to complement each other under occlusions. Public datasets and simulators including DAIR-V2X and OPV2V have enabled systematic evaluation across camera- and LiDAR-based stacks~\cite{Yu2022DAIRV2X,Xu2022OPV2V}. However, these approaches typically rely on CPM-style cooperative perception, which assumes access to rich sensor features or object reports. In contrast, the use of lightweight CAM-style object-level information to guide 3D object detectors remains much less explored.

Prior work has examined deep fusion of transmitted object boxes and demonstrated that localization noise can severely degrade their utility~\cite{Yuan2022_FPVRCNN}, while probabilistic frameworks highlight the importance of modeling uncertainty when combining CAM- or CPM-style object lists with onboard perception~\cite{Shan2022_ProbCP}. Building on these insights, we investigate how self-transmitted CAM-style information can be integrated into a BEV-based 3D detector and systematically quantify the sensitivity of this fusion to localization error, latency, and reduced V2X penetration.

\subsection{3D Object Detection and BEV Representations}
Modern 3D object detection increasingly relies on bird's-eye view (BEV) representations that align all sensor modalities in a common ground-plane coordinate system, providing a natural interface for multi-sensor fusion. While early monocular approaches~\cite{CHE16,DIN19} struggled with depth ambiguity, recent multi-view BEV detectors lift image features into a shared BEV space for unified geometric reasoning. Foundational methods such as \textit{BEVDepth}~\cite{li2022bevdepth}, and transformer-based designs like \textit{BEVFormer}~\cite{li2022bevformer} demonstrate the effectiveness of BEV feature lifting and temporal aggregation. \textit{BEVFusion}~\cite{bevfusion} exemplifies this paradigm by fusing camera, radar, and LiDAR in BEV through learned transformations, achieving state-of-the-art performance on nuScenes~\cite{caesar2020nuscenes}. However, existing BEV fusion methods do not exploit V2X object information, which provides complementary spatial coverage but arrives with localization errors and latency that can corrupt the BEV representation if naively integrated.

\section{Methodology}
\label{sec:method}

\begin{figure*}[t]
    \centering
    \includegraphics[width=1.0\textwidth]{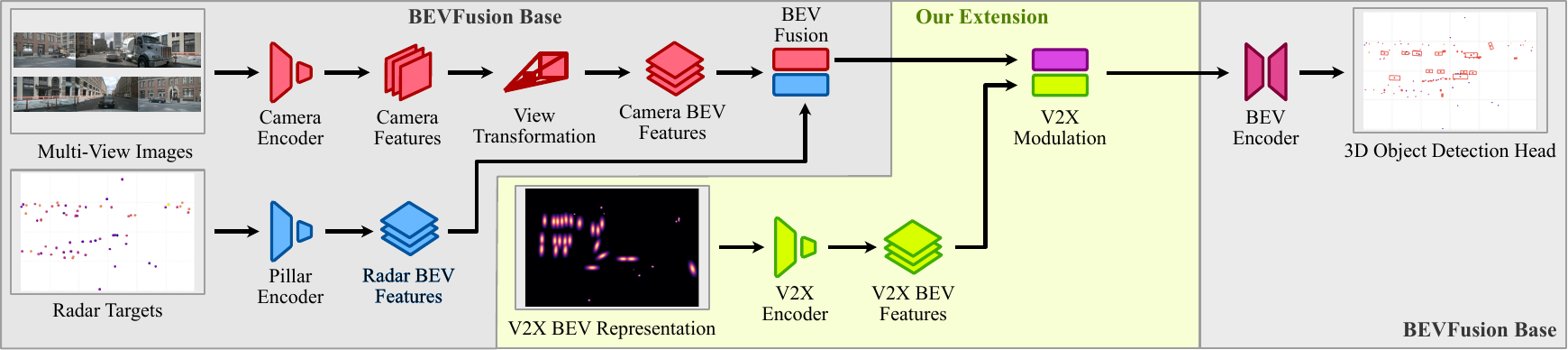}
    \caption{Architecture overview: \textit{BEVFusion}~\cite{bevfusion} with an added V2X branch and gating-based fusion at the BEV feature level. The V2X BEV input is encoded by a \textit{ResNet}~\cite{he2016deep} backbone, aligned with camera and radar features, and modulated by a learnable gate before entering the detection head to weight V2X contributions by reliability.}
    \label{fig:architecture}
\end{figure*}

Our goal is to integrate V2X object-level information into a \textit{BEVFusion}-style detector and to study robustness to V2X imperfections in a controlled way. To this end, we (A) emulate V2X messages from \textit{nuScenes} ground-truth annotations with configurable noise and object dropout, (B) convert the noisy object list into a multi-channel V2X BEV input, and (C) extend \textit{BEVFusion} with a V2X branch and gating-based fusion.

\subsection{V2X Emulation}
\label{subsec:v2x_emulation}
We implement ground-truth based V2X emulation on \textit{nuScenes}. For each frame, we use the 3D bounding box annotations (position, size, orientation, velocity, class) as a proxy for object-level V2X messages and inject controlled imperfections to mimic realistic behavior. Since \textit{nuScenes} annotates only sensor-visible objects, fully occluded actors are absent, making our evaluation conservative. This setup allows systematic variation of localization errors, latency effects, and missing senders. Importantly, we model both transmitter-side and ego-vehicle-side effects by separately applying object-level and frame-level perturbations. We apply four synthetic noise mechanisms:
\begin{itemize}
\item \textbf{Object-level rotation/translation noise:} Independent perturbations to each object’s heading and position.
\item \textbf{Frame-level rotation/translation noise:} Global perturbations applied uniformly to all objects in a frame.
\item \textbf{Object-level dropout:} Random removal of individual objects from the emulated V2X information.
\item \textbf{Class-level dropout:} Systematic removal of entire object classes from the emulated V2X information.
\end{itemize}
These perturbations correspond to common V2X failure modes: object-level noise reflects transmitter-side inaccuracies (e.g., GNSS error or stale states due to latency), frame-level noise models ego-pose drift affecting all received objects, object-level dropout represents missing transmitters, and class-level dropout captures non-broadcasting categories. In our studies, we sample the noise perturbation $\varepsilon$ from multiple distributions:
\begin{itemize}
\item \textbf{Uniform:} $\varepsilon \sim \mathcal{U}(a,b)$
\item \textbf{Gaussian (fixed $\sigma$):} $\varepsilon \sim \mathcal{N}(0,\sigma^2)$
\item \textbf{Gaussian (var. $\sigma$):} $\sigma \sim \mathcal{U}(\sigma_{\min},\sigma_{\max})$, $\varepsilon \sim \mathcal{N}(0,\sigma^2)$.
\end{itemize}

\begin{figure}[b]
    \centering
    \input{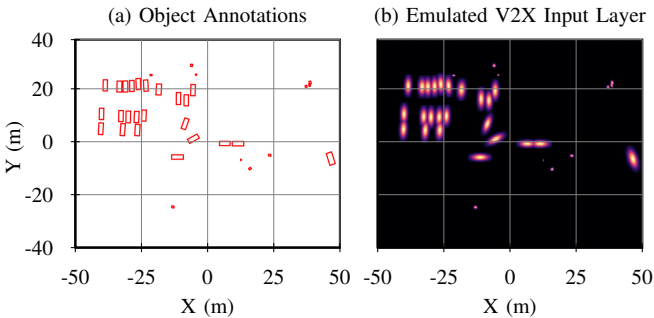}
    \vspace{-0.8cm}
    \caption{BEV input visualization: (left) ground-truth 3D boxes in the ego frame, (right) the corresponding Gaussian rasterized BEV map used as V2X input.}
    \label{fig:bev_visualization}
\end{figure}

\subsection{V2X BEV Input Generation}
\label{subsec:v2x_bev}
After applying noise and object dropout, we convert the resulting object list into a fixed-size BEV representation. We discretize the BEV plane into a grid and rasterize each object using an oriented anisotropic Gaussian footprint.

For a box with size $(d_x,d_y)$ and yaw $\theta$, we transform each relative offset $(\Delta x,\Delta y)$ from the box center to the box's local coordinate frame via
\[
u = \Delta x\cos\theta + \Delta y\sin\theta,\qquad
v = -\Delta x\sin\theta + \Delta y\cos\theta,
\]
then calculate an anisotropic Gaussian
\[
g(u,v) = \exp\!\left(-\tfrac{1}{2}\!\left[\left(\tfrac{u}{d_x/2}\right)^2 + \left(\tfrac{v}{d_y/2}\right)^2\right]\right).
\]

We write $g(u,v)$ into the corresponding class channel using a per-cell maximum over overlapping boxes. For additional channels encoding object properties (orientation, height, velocity), we multiply $g(u,v)$ by the normalized property value. The resulting BEV tensor contains
\begin{itemize}
\item one occupancy channel per object class;
\item two orientation channels: $\sin(\theta)$ and $\cos(\theta)$;
\item one height channel ($z$ coordinate);
\item and two velocity channels: $v_x$ and $v_y$.
\end{itemize}
All channels are normalized to the range $[0,1]$. A visualization of the BEV representation can be seen in Figure \ref{fig:bev_visualization}

Each object's contribution is weighted by a confidence value derived from the noise standard deviation. Formally, the confidence $c$ is an inverse function of the standard deviation $\sigma$. This mimics realistic V2X metadata (e.g., GNSS accuracy indicators). Crucially, we encode only the standard deviation used to generate noise, not the realized error, providing an uncertainty estimate rather than a correction signal. This encourages the network to learn appropriate trust levels for V2X information.

\subsection{Network Architecture and Fusion}
\label{subsec:architecture}
We build on \textit{BEVFusion}~\cite{bevfusion}, a multi-sensor fusion architecture that unifies sensor data in a shared BEV representation. This BEV-centric design is particularly suitable for V2X integration, as V2X messages can be converted into the same BEV space, enabling seamless, principled fusion with existing sensor modalities. We use only the camera and radar branches, disabling the original LiDAR branch. Figure~\ref{fig:architecture} provides an overview of the architecture and our V2X fusion scheme.

\subsubsection{Base \textit{BEVFusion} backbone}

A Swin-T backbone and LSS view transformer convert multi-view images into an $80$-channel BEV feature map, which is then passed to a BEV decoder and CenterHead detector. For radar+camera experiments, radar features are projected and fused following the standard \textit{BEVFusion} design. Additional configuration details are provided in Section~\ref{sec:setup}.

\subsubsection{V2X branch}
The V2X BEV input (Section~\ref{subsec:v2x_bev}) is processed by a dedicated branch: a \textit{ResNet} backbone and LSSFPN neck generate V2X BEV features that are spatially aligned with the camera and radar BEV representations, enabling fusion at a shared resolution.

\subsubsection{Gating-based fusion}
At \textit{BEVFusion}'s decoder stage, we inject V2X BEV features via a learnable gating mechanism. Let $\mathbf{F}_{\mathrm{fused}}$ denote the fused BEV features from camera and radar, and $\mathbf{F}_{\mathrm{v2x}}$ denote the encoded V2X features. A learnable gate produces spatially varying weights:
\[
\mathbf{G} = \sigma\!\left(\text{Conv}\left(\left[\mathbf{F}_{\mathrm{fused}}, \mathbf{F}_{\mathrm{v2x}}\right]\right)\right),
\]
where $\sigma$ is the sigmoid activation. The gated V2X features $\mathbf{G} \odot \mathbf{F}_{\mathrm{v2x}}$ are fused with the decoder features via learnable residual and scaling pathways. This mechanism enables adaptive weighting of V2X contributions based on the available fused features.

\subsubsection{Training noise and confidence}
During training, we apply Gaussian noise with uniformly sampled standard deviation per object, encoded as confidence in the V2X input. This exposes the network to varying V2X confidence levels. Frame-level noise is applied the same way and contributes to the confidence of all objects simultaneously. At test time, we evaluate under various noise and object dropout configurations using Gaussian noise with fixed standard deviation to characterize robustness. For exact configurations, see Section~\ref{sec:setup}.

\subsubsection{V2X dropout for V2X penetration rate modeling}
To emulate V2X penetration rates, we apply object-level V2X dropout. During training, this is implemented as Bernoulli dropout on individual V2X objects with fixed drop rates (\SI{25}{\percent}, \SI{50}{\percent}, \SI{75}{\percent}), corresponding to different assumed V2X penetration rates. At test time, we sweep the drop rate for a given model to obtain performance as a function of the effective V2X penetration rate, as detailed in Section~\ref{sec:setup}. This setup allows us to study how training with missing V2X data affects robustness to reduced V2X penetration.

\section{Experimental Setup and Evaluation Methodology}
\label{sec:setup}
This section details the dataset, metrics, model configurations, training setup, noise models, and ablation studies used to evaluate V2X integration.

\subsection{Experimental Overview}

\subsubsection{Metrics}
We follow the official \textit{nuScenes} evaluation protocol~\cite{caesar2020nuscenes}. Unless stated otherwise, \textit{nuScenes} Detection Score (NDS) is computed over all ten object categories. For class-wise analyses (Section~\ref{ssec:coverage}), we additionally report per-class average precision (AP).

\subsubsection{Dataset}
All experiments use \textit{nuScenes} with the official train/validation/test splits and 10-class taxonomy~\cite{caesar2020nuscenes}. V2X emulation follows Section~\ref{subsec:v2x_emulation}. Because \textit{nuScenes} annotations only cover objects visible to the onboard perception sensors, fully occluded actors are absent. This makes our V2X gains conservative, as real deployments can surface such occluded participants.

\subsubsection{Model configuration}
Input images are resized to $256{\times}704$ pixels. The BEV grid spans \SIrange{-51.2}{51.2}{\meter} in $x$ and $y$ at \SI{0.4}{\meter} resolution, with a depth range of \SIrange{1}{60}{\meter} and vertical extent \SIrange{-10}{10}{\meter}. The view transformer yields an $80$-channel BEV feature map processed by a BEV decoder and a CenterHead with output stride~8.

\subsubsection{Modalities}
We consider two sensor configurations:
\begin{itemize}
    \item \textbf{Camera-only \textit{BEVFusion}:} 6 multi-view cameras.
    \item \textbf{Radar+Camera \textit{BEVFusion}:} fusion of the front-center radar and front-center camera.
\end{itemize}
For both configurations, we train baselines without V2X and variants with the V2X branch and gating-based fusion.

\subsubsection{Training}
We train for 20~epochs using AdamW with learning rate $2\times10^{-4}$, weight decay $0.01$, and a cyclic learning rate schedule with a 500-iteration warm-up, using FP16. Data augmentation follows the standard \textit{BEVFusion} recipe, including 2D resizing and rotation, 3D rotation and scaling, and horizontal flips. V2X-augmented networks use the noise and confidence encoding described in Section~\ref{subsec:architecture}.

\subsection{Noise Configurations}

We assess robustness using single-parameter noise sweeps and combined sweeps applied to emulated V2X data. The $0.3$\,rad and $2$\,m maxima reflect common GNSS/odometry drift and short-latency pose errors; lower steps cover moderate misalignment. Combined levels mirror these ranges so joint perturbations stay realistic while spanning mild to severe degradation.

\subsubsection{Single-parameter sweeps}
We vary one noise source at a time while keeping all others at zero. We independently sweep object rotation, object translation, frame rotation, and frame translation in the $x$-$y$ plane. Rotation sweeps use $\sigma_{\text{rot}}\in\{0.0, 0.05, 0.1, 0.15, 0.2, 0.3\}\,\si{\radian}$, and translation sweeps use $\sigma_{\text{trans}}\in\{0.0, 0.25, 0.5, 0.75, 1.0, 1.5, 2.0\}\,\si{\meter}$.

\subsubsection{Combined sweeps}
All noise types are simultaneously active using the configurations in Table~\ref{tab:combined_sweeps}-a.

\subsubsection{Training noise levels}
The networks in Section~\ref{sec:setup:ablations} use uniformly sampled Gaussian noise during training with the ranges in Table~\ref{tab:train_noise}-b.

\begin{table}[t]
\centering
\caption{V2X Noise Configurations}
\label{tab:noise_configs}

\vspace{-0.3em}
\noindent(a) Noise during test: fixed $\sigma$
\label{tab:combined_sweeps}
\vspace{0.5em}

\noindent\begin{tabular}{lcccc}
\hline
\textbf{Noise Level} & \multicolumn{2}{c}{\textbf{Object Level}} & \multicolumn{2}{c}{\textbf{Frame Level}} \\
 & \textbf{Rot. [\si{\radian}]} & \textbf{Trans. [\si{\meter}]} & \textbf{Rot. [\si{\radian}]} & \textbf{Trans. [\si{\meter}]} \\
\hline
Low & 0.05 & 0.25 & 0.05 & 0.25 \\
Medium-low & 0.10 & 0.50 & 0.10 & 0.50 \\
Medium-high & 0.15 & 0.75 & 0.15 & 0.75 \\
High & 0.30 & 1.50 & 0.30 & 1.50 \\
\hline
\end{tabular}

\vspace{1em}
\noindent(b) Noise during training: $\sigma$ sampled from range
\label{tab:train_noise}
\vspace{0.5em}

\noindent\begin{tabular}{lcccc}
\hline
\textbf{Level} & \multicolumn{2}{c}{\textbf{Object Level}} & \multicolumn{2}{c}{\textbf{Frame Level}} \\
 & \textbf{Rot. [\si{\radian}]} & \textbf{Trans. [\si{\meter}]} & \textbf{Rot. [\si{\radian}]} & \textbf{Trans. [\si{\meter}]} \\
\hline
Low & 0.0005--0.005 & 0.01--1.0 & 0.0005--0.005 & 0.01--1.0 \\
Medium & 0.001--0.01 & 0.02--2.0 & 0.001--0.01 & 0.02--2.0 \\
High & 0.002--0.02 & 0.03--3.0 & 0.002--0.02 & 0.03--3.0 \\
\hline
\end{tabular}
\end{table}

\subsection{Ablation Studies}
\label{sec:setup:ablations}
We perform five experiments (Experiments~A--E), all using the training and evaluation protocol defined above.

\subsubsection*{Experiment~A: V2X without training noise}
We train a camera+V2X model without V2X noise to measure its best-case benefit and sensitivity to noise.

\subsubsection*{Experiment~B: Training with noise}
We train camera+V2X and radar+camera+V2X \textit{BEVFusion} models with noise-enabled V2X. Gaussian noise with uniformly sampled standard deviation is injected and encoded as confidence, encouraging confidence-aware fusion for noisy V2X inputs.

\subsubsection*{Experiment~C: Training noise level}
We train low-, \mbox{medium-,} and high-noise variants for each configuration to study how the magnitude of training-time V2X noise influences the learned fusion behavior. This ablation isolates whether exposure to noise during training causes the model to rely on V2X inputs, learn confidence-aware fusion that is robust to degradation, or downweight noisy V2X signals.

\subsubsection*{Experiment~D: Class-limited V2X Input}
\label{ssec:coverage}
We restrict V2X information to classes likely to carry V2X units (car, truck, bus, trailer, motorcycle, and bicycle), explicitly excluding pedestrians, cones, and barriers. This setup enables class-wise AP analysis to assess how different semantic categories are affected by the availability of V2X information. For brevity, we report only camera+V2X results. 

\subsubsection*{Experiment~E: V2X penetration rate}
We simulate varying V2X penetration rates by randomly dropping object-level V2X messages at test time for the model from Experiment~D. The penetration rate is defined as the fraction of V2X-capable actors that transmit messages. To test whether training with missing messages improves robustness, we also train three camera+V2X variants with fixed training-time dropout rates of \SI{25}{\percent}, \SI{50}{\percent}, and \SI{75}{\percent}. At evaluation, we measure performance for the original and dropout-trained models across test-time penetration rates of \SI{25}{\percent}, \SI{50}{\percent}, \SI{75}{\percent}, and \SI{100}{\percent}. This experiment assesses whether exposure to object dropout during training reduces sensitivity to reduced V2X penetration.

\section{Results}
\label{sec:results}

In this section, we present the evaluation of our V2X integration approach, following the ablation studies defined in Section~\ref{sec:setup:ablations}. We analyze the potential of V2X integration, the impact of noise-aware training on robustness, the trade-off between peak performance and stability, and the effects of class-limited V2X input and V2X penetration rates.

\begin{figure*}[t]
    \centering
    \input{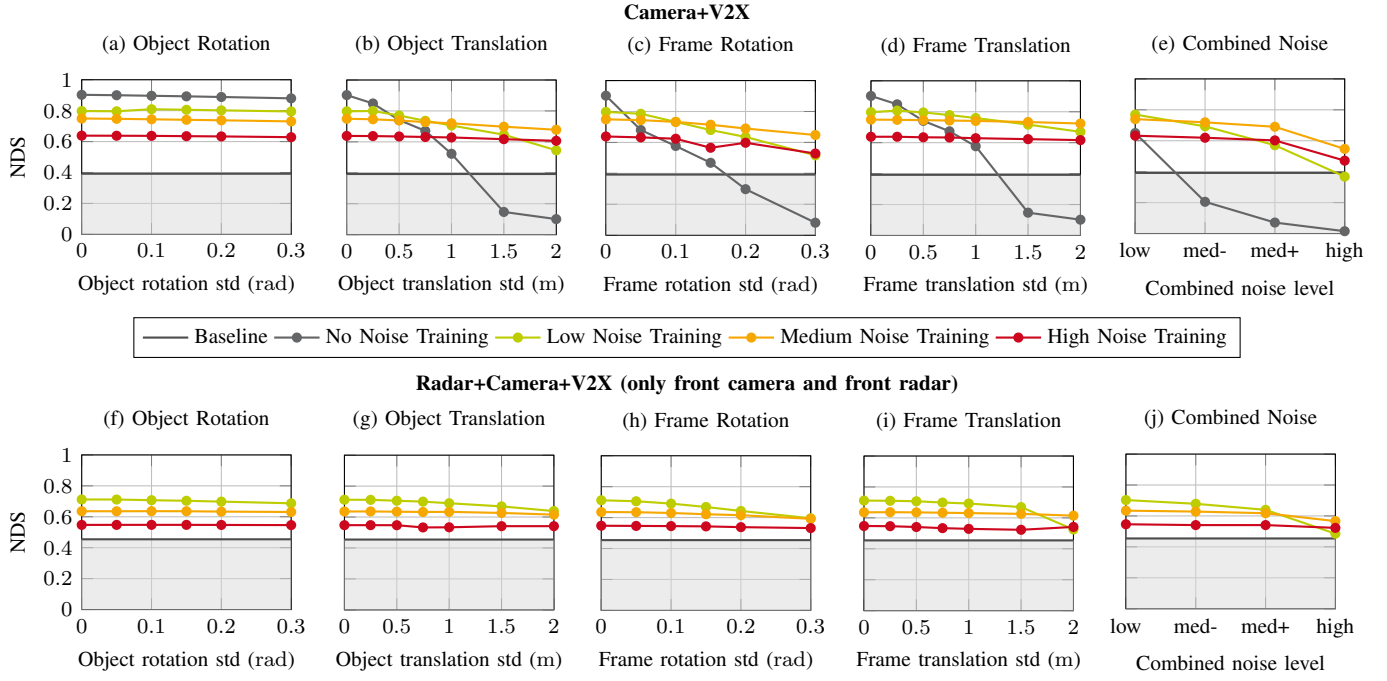}
    \caption{Noise robustness for Experiments~A--C: NDS versus noise for camera+V2X (top, panels (a)--(e)) and radar+camera+V2X (bottom, panels (f)--(j)). Columns sweep object/frame rotation and translation plus a combined setting from Table~\ref{tab:combined_sweeps}.}
    \label{fig:noise_robustness}
\end{figure*}

\subsection*{Experiment A: Potential of V2X Integration}
To assess the upper bound of performance gains from V2X integration, we first evaluate a camera+V2X \textit{BEVFusion} model trained with ideal, noise-free V2X input.
Under perfect conditions (no test-time noise), this model achieves an NDS of $0.90$, a substantial improvement over the camera-only baseline (NDS $0.39$). This large gain is expected, as noise-free V2X emulation effectively provides ground-truth object information, drastically enhancing perception by resolving occlusions and providing accurate depth and velocity cues.

However, as illustrated in Figure~\ref{fig:noise_robustness}, this performance is highly fragile. While the model is relatively robust to object-level rotation noise (Figure~\ref{fig:noise_robustness}a), it exhibits catastrophic degradation under other perturbations. For instance, frame rotation (Figure~\ref{fig:noise_robustness}c) causes NDS to drop to $0.08$, and translation errors (Figure~\ref{fig:noise_robustness}b, d) lead to NDS values around $0.10$ at $\sigma=2\,\si{\meter}$. This indicates that without exposure to noise during training, the network over-relies on precise spatial alignment and fails to generalize to realistic V2X imperfections.

\subsection*{Experiment B: Robustness through Noise-Aware Training}
Training with Gaussian V2X noise substantially improves robustness. In the combined noise setting (Figure~\ref{fig:noise_robustness}e), which simultaneously applies object- and frame-level rotation and translation noise (Table~\ref{tab:combined_sweeps}), models trained with low, medium, or high noise degrade gracefully and stay close to or above the camera-only baseline across most settings. In stark contrast, the zero-noise model from Experiment~A collapses to NDS~$0.02$ at high combined noise. This demonstrates that noise-aware training enables the network to learn confidence-weighted fusion, effectively down-weighting V2X information when it is unreliable rather than over-relying on it.

To provide context for alternative sensor modalities, we also evaluate radar+camera fusion (front radar and front camera) under identical noise conditions. Figure~\ref{fig:noise_robustness}j shows that radar+camera models trained with low and medium noise levels follow qualitatively similar robustness trends as camera+V2X and maintain performance well above the radar+camera baseline across all noise settings. Under maximum combined noise, the low-noise radar+camera model still achieves NDS~$0.49$, illustrating the inherent robustness benefit of adding radar to the fusion stack.

\subsection*{Experiment C: Impact of Training Noise Magnitude}
Varying the training noise level reveals a trade-off between peak performance and robustness. For camera+V2X (Figure~\ref{fig:noise_robustness}e), the low-noise trained model peaks highest with NDS~$0.80$) but collapses under heavy test noise to NDS~$0.37$, even below the camera-only baseline ($0.39$). The high-noise trained model starts lower ($0.64$) yet remains stable, dropping only to $0.53$. The medium-noise trained model strikes a balance, retaining most clean-scenario accuracy while staying consistent across noise sweeps. This shows that higher training noise mitigates V2X over-reliance at the expense of peak performance.

Radar+camera fusion follows the same pattern. The low-noise trained model peaks at NDS~$0.71$ but falls to $0.49$ under high test noise, whereas the high-noise trained model peaks at $0.55$ and degrades only to $0.52$. The medium-noise variant again balances the trade-off, starting at $0.64$ and remaining steady at $0.57$. Overall, Figure~\ref{fig:noise_robustness} demonstrates that noise-aware training induces a largely modality-agnostic robustness–accuracy trade-off.

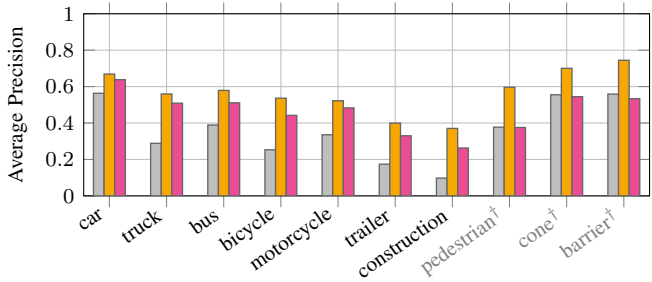
\begin{figure}[!b]
    \centering
    \begin{tikzpicture}
\begin{axis}[
    ybar=0pt,
    bar width=4pt,
    width=0.50\textwidth,
    height=0.22\textwidth,
    font=\footnotesize,
    ylabel={Average Precision},
    symbolic x coords={car, truck, bus, bicycle, motorcycle, trailer, construction, pedestrian, cone, barrier},
    xtick=data,
    xticklabel style={rotate=35,anchor=east,font=\footnotesize},
    xticklabels={car, truck, bus, bicycle, motorcycle, trailer, construction, {\color{gray}pedestrian$^\dagger$}, {\color{gray}cone$^\dagger$}, {\color{gray}barrier$^\dagger$}},
    ymin=0, ymax=1.0,
    ymajorgrids,
    xmajorgrids,
    enlarge x limits=0.05,
    legend pos=north west]
    
    \addplot[ybar,fill=gray!50,draw=rwthgrey,line width=0.5pt] coordinates
        {(car,0.5635) (truck,0.2885) (bus,0.3892) (bicycle,0.2532) (motorcycle,0.3355) (trailer,0.1740) (construction,0.0975) (pedestrian,0.3772) (cone,0.5551) (barrier,0.5588)};

    \addplot[ybar,fill=rwthorange,draw=rwthgrey,line width=0.5pt] coordinates
        {(car,0.6687) (truck,0.5596) (bus,0.5791) (bicycle,0.5364) (motorcycle,0.5224) (trailer,0.3996) (construction,0.3703) (pedestrian,0.5965) (cone,0.7004) (barrier,0.7445)};

    \addplot[ybar,fill=rwthmagenta!70,draw=rwthgrey,line width=0.5pt] coordinates
    {(car,0.6381) (truck,0.5090) (bus,0.5108) (bicycle,0.4421) (motorcycle,0.4831) (trailer,0.3300) (construction,0.2629) (pedestrian,0.3754) (cone,0.5446) (barrier,0.5338)};
\end{axis}
\end{tikzpicture}
    \caption{Per-class AP under medium-high noise comparing: camera-only baseline\protect\colorcircle{gray!50}, camera+V2X with full V2X input (medium training noise)\protect\colorcircle{rwthorange}, and camera+V2X restricted to transmitting classes only\protect\colorcircle{rwthmagenta!70}. Classes marked with $\dagger$ are non-transmitting and excluded from V2X input. All V2X models are trained with medium noise and evaluated at medium-high noise.}
    \label{fig:class_wise_map}
\end{figure}

\subsection*{Experiment D: Effect of Class-Limited V2X Input}
This experiment isolates the impact of restricting V2X to actively transmitting classes by comparing two identically trained camera+V2X models: one with full V2X coverage and one with V2X disabled for pedestrians, cones, and barriers. Both networks were trained with medium noise (Table \ref{tab:train_noise}-a) and evaluated under medium-high noise (Table \ref{tab:combined_sweeps}-b). Results reflect a single training run, so class-level metrics may vary due to initialization and sampling noise.
The per-class analysis in Figure~\ref{fig:class_wise_map} shows that V2X availability influences classes asymmetrically. Transmitting classes continue to exhibit strong gains over the baseline. For example, cars achieve an AP of 0.64 compared to the 0.56 baseline, and motorcycles also see substantial improvements. Non-transmitting classes align more closely with their camera-only baselines, with pedestrian AP at 0.38 (baseline 0.37) and traffic cones at 0.54 (baseline 0.55). This pattern indicates that the network exploits V2X cues where present while maintaining baseline-level camera performance for classes without V2X support.

\subsection*{Experiment E: Influence of V2X Penetration Rate}
\begin{figure}[t]
    \centering
    \begin{tikzpicture}
    \begin{axis}[
        width=0.48\textwidth,
        height=0.3\textwidth,
        xlabel={V2X penetration rate},
        ylabel={NDS},
        xmin=0.25, xmax=1.0,
        ymin=0.0, ymax=0.8,
        xtick={0.25, 0.5, 0.75, 1.0},
        ymajorgrids,
        xmajorgrids,
        legend pos=south east,
        legend style={font=\footnotesize},
        font=\footnotesize,
        title={Object Dropout Robustness}]
        
        \addplot[fill=gray!30,fill opacity=0.5,draw=none,forget plot] coordinates
            {(0.25,0) (1.0,0) (1.0,0.3921) (0.25,0.3921)};
        \addplot[color=black!70,thick] coordinates {(0.25,0.3921) (1.0,0.3921)};
        \addlegendentry{Baseline}

        \addplot[mark=*,color=rwthorange,thick,mark size=1.5pt] coordinates
            {(0.25,0.3433) (0.5,0.4601) (0.75,0.6297) (1.0,0.6907)};
        \addlegendentry{No object dropout (med. noise)}

        \addplot[mark=*,color=rwthmaygreen,thick,mark size=1.5pt] coordinates
            {(0.25,0.4357) (0.5,0.4755) (0.75,0.5478) (1.0,0.6156)};
        \addlegendentry{25\% object dropout (train)}

        \addplot[mark=*,color=rwthblue,thick,mark size=1.5pt] coordinates
            {(0.25,0.4300) (0.5,0.4582) (0.75,0.5232) (1.0,0.5425)};
        \addlegendentry{50\% object dropout (train)}

        \addplot[mark=*,color=rwthmagenta,thick,mark size=1.5pt] coordinates
            {(0.25,0.4049) (0.5,0.4259) (0.75,0.4771) (1.0,0.4925)};
        \addlegendentry{75\% object dropout (train)}
        
    \end{axis}
\end{tikzpicture}
    \caption{Robustness to partial V2X penetration: NDS versus V2X penetration rate. The no-dropout V2X model peaks at full penetration but falls steeply as availability drops, while dropout-trained variants trade peak NDS for flatter performance across penetration rates.}
    \label{fig:dropout_robustness}
\end{figure}
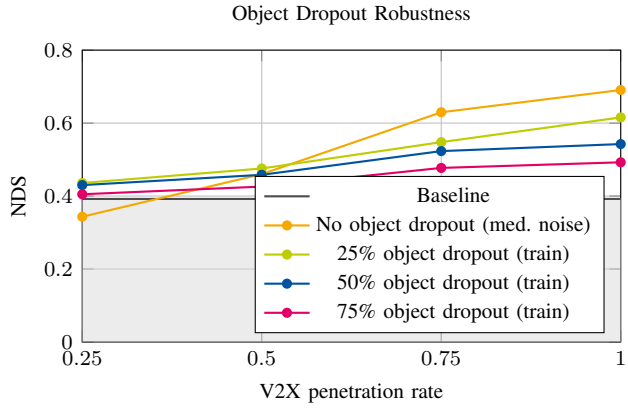

Figure~\ref{fig:dropout_robustness} shows how performance varies with V2X penetration rate. The model from Experiment~D, trained without object dropout (orange), performs well under full penetration (NDS~$0.69$ at \SI{100}{\percent}~penetration) but degrades sharply as penetration decreases, reaching NDS~$0.34$ at \SI{25}{\percent}~penetration. Models trained with V2X dropout exhibit substantially improved robustness. The low-dropout variant (\SI{25}{\percent}, green) achieves NDS~$0.44$ at \SI{25}{\percent}~penetration, and the medium and high-dropout models (\SI{50}{\percent}, \SI{75}{\percent}) further flatten the performance curve at the cost of a reduced peak NDS. Overall, incorporating object dropout during training yields a much more stable performance profile across partial V2X penetration scenarios, indicating that robustness to missing V2X messages is critical during real-world rollout.

\FloatBarrier

\section{Discussion and Conclusion}
Our study shows that object-level V2X, where each participant broadcasts information about its own state, can meaningfully improve a \textit{BEVFusion}-style detector, but only when fusion explicitly accounts for V2X uncertainty and intermittent availability. Because \textit{nuScenes} annotates only sensor-visible objects, our results are conservative relative to deployments where V2X can reveal fully occluded actors.

Experiments~A-C highlight that V2X must be treated as an uncertainty-bearing input rather than ground truth. Training on perfectly clean V2X induces brittle over-reliance, whereas noise-aware training with confidence cues enables adaptive gating that balances V2X against ego-sensor evidence. The injected noise level defines an operating point: low noise maximizes clean-condition NDS, while higher noise trades a small performance drop for substantially improved robustness under spatial perturbations.

Experiments~D and~E show that V2X benefits depend on both message quality and V2X penetration rates. Even at low penetration, models trained with dropout retain meaningful improvements over the baseline, and gains grow with penetration. This underscores that early deployments can already deliver relevant benefits, while higher coverage amplifies them.

Our emulation assumes independent Gaussian pose errors and does not capture correlated failures, non-Gaussian noise, time-varying latency, or feature-level V2X messages. The absence of fully occluded actors in \textit{nuScenes} bounds our evaluation; datasets with occlusion-heavy scenes and real V2X traces are essential to quantify the full potential of cooperative awareness data in 3D object detection. Future work should incorporate temporal and correlated error models, online uncertainty calibration, and stress-testing on occlusion-aware benchmarks.

In summary, our noise-aware V2X fusion improves 3D object detection across a wide range of V2X quality and penetration rates, with graceful degradation as V2X deteriorates.

\bibliographystyle{IEEEtran}
\bibliography{refs}

\end{document}